\pdfoutput=1

\documentclass[11pt]{article}

\usepackage[]{acl}

\usepackage{times}
\usepackage{latexsym}

\usepackage[T1]{fontenc}

\usepackage[utf8]{inputenc}
\colorlet{shadecolor}{gray!40}

\usepackage{microtype}

\usepackage{amsmath,amsfonts,bm}

\def\eqref#1{equation~\ref{#1}}

\def\1{\bm{1}}

\DeclareMathAlphabet{\mathsfit}{\encodingdefault}{\sfdefault}{m}{sl}
\SetMathAlphabet{\mathsfit}{bold}{\encodingdefault}{\sfdefault}{bx}{n}

\usepackage[T2A]{fontenc}
\usepackage{xcolor,colortbl}
\usepackage{arydshln}
\usepackage{hyperref}
\usepackage{tcolorbox}
\usepackage{caption}
\usepackage{subcaption}
\usepackage{placeins}
\usepackage{arydshln}
\usepackage{CJKutf8}
\usepackage{multirow}
\usepackage{pythonhighlight}
\usepackage[T2A,T1]{fontenc}
\usepackage{ulem}
\usepackage{booktabs}
\usepackage{pifont}
\usepackage{listings}
\usepackage[colorinlistoftodos,prependcaption]{todonotes}

\newcommand{\cmark}{\ding{51}}%
\newcommand{\xmark}{\ding{55}}%

\definecolor{codegreen}{rgb}{0,0.6,0}
\definecolor{codegray}{rgb}{0.5,0.5,0.5}
\definecolor{codepurple}{rgb}{0.58,0,0.82}
\definecolor{backcolour}{rgb}{0.95,0.95,0.92}

\lstdefinestyle{mystyle}{
    backgroundcolor=\color{backcolour},   
    commentstyle=\color{codegreen},
    keywordstyle=\color{magenta},
    numberstyle=\tiny\color{codegray},
    stringstyle=\color{codepurple},
    basicstyle=\ttfamily\footnotesize,
    breakatwhitespace=false,         
    breaklines=true,                 
    captionpos=b,                    
    keepspaces=true,                 
    numbers=left,                    
    numbersep=5pt,                  
    showspaces=false,                
    showstringspaces=false,
    showtabs=false,                  
    tabsize=2
}

\definecolor{lightgray}{rgb}{.9,.9,.9}
\definecolor{darkgray}{rgb}{.4,.4,.4}
\definecolor{purple}{rgb}{0.65, 0.12, 0.82}
\lstdefinelanguage{JavaScript}{
  keywords={break, case, catch, continue, debugger, default, delete, do, else, false, finally, for, function, if, in, instanceof, new, null, return, switch, this, throw, true, try, typeof, var, void, while, with},
  morecomment=[l]{//},
  morecomment=[s]{/*}{*/},
  morestring=[b]',
  morestring=[b]",
  ndkeywords={class, export, boolean, throw, implements, import, this}
}
\usepackage{hyperref}
\usepackage{url}

\title{\texttt{ICE-Score}: Instructing Large Language Models to Evaluate Code}

\author{Terry Yue Zhuo \\
Monash University and CSIRO's Data61 \\
\texttt{terry.zhuo@monash.edu}
}

\begin{document}
\maketitle
\begin{abstract}
Recent advancements in the field of natural language generation have facilitated the use of large language models to assess the quality of generated text. Although these models have shown promising results in tasks such as machine translation and summarization, their applicability in code intelligence tasks remains limited without human involvement. The complexity of programming concepts required for such tasks makes it difficult to develop evaluation metrics that align with human judgment. Token-matching-based metrics, such as BLEU, have demonstrated weak correlations with human practitioners in code intelligence tasks. Moreover, utilizing human-written test suites to evaluate functional correctness can be challenging in domains with low resources. To overcome these obstacles, we propose \texttt{ICE-Score}, a new evaluation metric via instructing large language models (LLMs) for code assessments. Our metric addresses the limitations of existing approaches by achieving superior correlations with functional correctness and human preferences, without the need for test oracles or references. We evaluate the efficacy of our metric on two different aspects (\textit{human preference} and \textit{execution success}) and four programming languages. Our results demonstrate that our metric surpasses state-of-the-art metrics for code generation, delivering high levels of accuracy and consistency across various programming languages and tasks. We also make our evaluation metric and datasets available to the public\footnote{\url{https://github.com/terryyz/ice-score}}, encouraging further research in evaluating code intelligence tasks.
\end{abstract}

\section{Introduction}

Natural language generation (NLG) systems have seen significant progress in developing large language models (LLMs). These models have shown great promise in generating high-quality and diverse texts that can be difficult to distinguish from human-written texts~\citep{ouyang2022training}. However, evaluating the quality of NLG systems remains a challenging task, primarily due to the limitations of traditional evaluation metrics. Token-matching-based metrics, such as BLEU~\citep{papineni2002bleu} and ROUGE~\citep{lin2004rouge}, have been widely used to evaluate NLG systems but have demonstrated poor correlation with human judgment and a lack of ability to capture semantic meanings~\citep{kocmi-etal-2021-ship}. Furthermore, these metrics require reference output, which can be challenging to obtain for new tasks and low-resource domains~\citep{liu2023gpteval}.

In recent years, the use of LLMs as reference-free evaluators for Natural Language Generation (NLG) tasks has gained attention among researchers. This approach is strongly aligned with human preferences, even when reference texts are 
unavailable~\citep{liu2023gpteval,fu2023gptscore}. The underlying assumption behind this approach is that LLMs possess a profound understanding of human-generated text and task instructions, enabling them to evaluate various NLG tasks through prompts. The exceptional performance of LLMs in contextual understanding and natural language generation, as evidenced by studies~\citep{brown2020language}, further supports this assumption. Moreover, LLMs trained on both textual and code-based data have showcased remarkable capabilities in diverse downstream tasks related to source code, including code generation~\citep{openai2023gpt4,allal2023santacoder,li2023starcoder}. While a performance gap still exists between LLMs and human developers in code-related tasks, recent research has illustrated that LLMs can be enhanced to handle various source code tasks with appropriate guidance~\citep{chen2023teaching,madaan2023self}. This indicates the significant potential of LLMs in comprehending and working with source code.

\begin{figure*}
    \centering
    \resizebox{\textwidth}{!}{\includegraphics{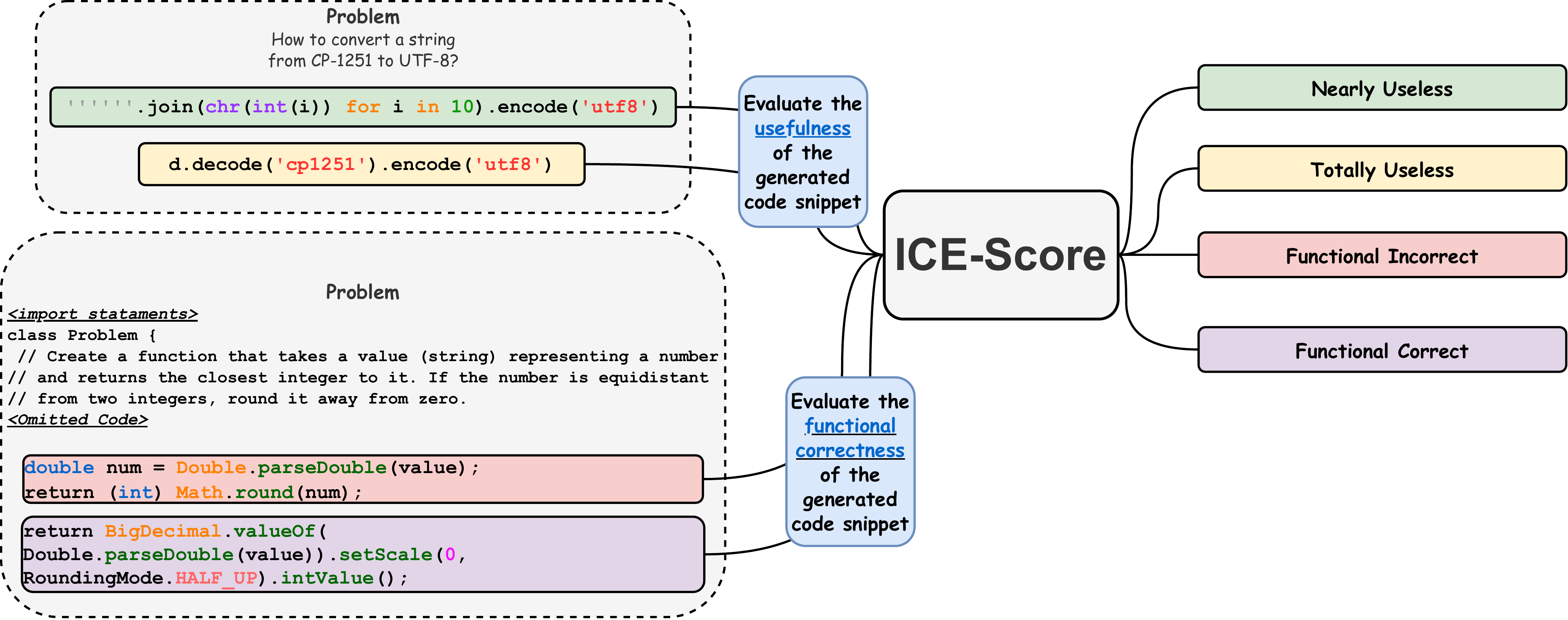}}
    \caption{An illustration of \texttt{ICE-Score}. On the left-hand side, we input the task problems and corresponding generated code snippets. On the right-hand side, \texttt{ICE-Score} outputs the corresponding assessments.}
    \label{fig:eval_overview}
\end{figure*}
Code evaluation presents unique challenges, requiring a deeper understanding of programming concepts and more complex syntax than natural language generation~\citep{hindle2016naturalness}. Traditional reference-based evaluation metrics for code generation, such as BLEU~\cite{papineni2002bleu}, ROUGE~\citep{lin2004rouge}, and chrF~\citep{popovic2015chrf}, rely on token matching to assess performance automatically. However, these metrics have demonstrated poor correlation with human evaluation~\citep{evtikhiev2022bleu} since they often underestimate the variety of outputs with the same semantic logic. While some studies have incorporated programming features to improve these metrics, they have shown limited gains and poor correlation with functional correctness~\citep{eghbali2022crystalbleu,tran2019does}. Alternatively, researchers have proposed using well-designed test suites to objectively evaluate code generation performance at the function level~\citep{chen2021evaluating, zheng2023codegeex,cassano:multipl-e}. However, developing these test suites requires programming expertise, which can be impractical and costly in low-resource scenarios. Additionally, executing model-generated code poses a security risk and must be run in an isolated sandbox, which is technically cumbersome.

More recently, CodeBERTScore~\citep{zhou2023codebertscore}, a neural-model-based evaluation metric, has been proposed, showing a higher correlation with functional correctness and human preferences by capturing the semantic information of reference code and generated code. However, CodeBERTScore still relies on high-quality references that can be difficult and expensive to obtain. Moreover, the limited performance of the CodeBERT~\citep{feng2020codebert} backbone suggests that it has not yet reached a human-level understanding of source code, limiting the effectiveness of CodeBERTScore. Therefore, more advanced evaluation metrics are needed so that they can better capture the complex syntax and semantics of code intelligence tasks.

To address these challenges, we propose a novel evaluation metric based on LLMs trained on both text and code, shown in Figure~\ref{fig:eval_overview}. Specifically, we \underline{I}nstruct LLMs to perform human-like multi-dimensional \underline{C}ode \underline{E}valuation, where the metric is denoted as \texttt{ICE-Score}. Our metric leverages the recent NLG metric, G-EVAL~\citep{liu2023gpteval}, but achieves superior correlations with subjective human preferences and objective functional correctness, both at the example and corpus levels. Different from G-EVAL, \texttt{ICE-Score} only relies on assessment criteria and evaluation step template, without the need for instruction generation and weighted scoring function.

Based on our extensive evaluation, we have summarized our contributions as follows:

\begin{itemize}
    \item We designed the first multi-dimensional and reference-free automatic evaluation metric for code intelligence tasks via large language models.
    \item We conducted extensive experiments to demonstrate the efficacy of \texttt{ICE-Score} on four programming languages (Java, Python, C, C++, and JavaScript) from two aspects (\textit{human-based usefulness} and \textit{execution-based functional correctness}). 
    \item We further discussed several aspects that can improve the performance of \texttt{ICE-Score}, including the backbone model performance and reasoning capability.
\end{itemize}

\section{Method}
Our evaluation metric \texttt{ICE-Score}, inspired by G-EVAL~\cite{liu2023gpteval}, consists of two main components: 1) task definition, evaluation criteria, and detailed evaluation steps, and 2) a given problem and generated code snippet for evaluation. Different from G-EVAL, we only require the input of evaluation criteria and template-based evaluation steps, without the need for generation from LLMs. In addition, As we set the model temperature to 0, our evaluation metric no longer needs a weighted scoring function after iterative score generation. These two differences suggest that \texttt{ICE-Score} is more cost-friendly and efficient.

\subsection{Instructions for Code Evaluation} 
The evaluation of code quality involves two main aspects: 1) human judgment of code usefulness and 2) execution-based functional correctness. To provide a comprehensive evaluation, we adopt the design of G-EVAL for the general task instruction, as follows:
\begin{quote}
\textit{You will be given the code snippet for a problem. 
Your task is to rate the code snippet only on one metric. Please make sure you read and understand these instructions carefully. Please keep this document open while reviewing, and refer to it as needed.}
\end{quote}

Regarding the task-agnostic prompt, we have designed the evaluation criteria for assessing \textbf{code usefulness}, as shown in Appendix~\ref{appendix:usefulness}. These criteria are aligned with previous human evaluations of code quality~\citep{evtikhiev2022bleu}. To evaluate \textbf{functional correctness}, we emphasize the importance of considering unit tests during the evaluation process. We present the following criteria for evaluating functional correctness, as provided in Appendix~\ref{appendix:correctness}.

For the instruction of evaluation steps, we provide a template-based prompt:
\begin{quote}
\textit{Evaluation Steps:}\\
\textit{1. Read the problem carefully and identify the required functionalities of the implementation.}\\
\textit{2. Read the code snippet and compare it to the problem. Check if the code snippet covers all required functionalities of the problem, and if it aligns with the Evaluation Criteria.}\\
\textit{3. Assign a score for \textbf{[Evaluation Aspect]} on a scale of 0 to 4, where 0 is the lowest and 4 is the highest based on the Evaluation Criteria.}\\
\end{quote}

Here, we define \textbf{[Evaluation Aspect]} as any aspects that are emphasized during the evaluation. In our paper, we consider \textbf{code usefulness} and \textbf{functional correctness}.
\subsection{Inputs of Code Evaluation} 
It is worth noting that most code generative models do not take formatting into account, resulting in unformatted code that requires post-processing of code formatting to be understood, compiled, and executed~\citep{zheng2023codegeex}. Additionally, automatic evaluation metrics for code generation, such as CodeBLEU~\citep{ren2020codebleu} and RUBY~\citep{tran2019does}, still rely on language-specific program parsers~\footnote{\url{https://tree-sitter.github.io/}}. However, based on prior findings that LLMs can robustly understand input data~\citep{huang2022large,zhuo2023red,zhu2023promptbench}, we hypothesize that LLMs can also understand programming context without proper formatting. Therefore, for evaluation, we input the problems and generated code (and reference code, if provided). When the reference code is provided, we slightly modify the evaluation steps in the prompt to incorporate it.

\section{Experiment Setup}
We evaluate the effectiveness of \texttt{ICE-Score} using GPT-3.5 (\texttt{GPT-3.5-turbo}\footnote{\url{https://platform.openai.com/docs/models/gpt-3-5}}) as the backbone across multiple datasets and programming languages. We conduct two experiments to investigate the correlation between \texttt{ICE-Score} and human preference and functional correctness, respectively. We compare the performance of LLM-based evaluations against 7 predominant automatic evaluation metrics, including the state-of-the-art CodeBERTScore~\citep{zhou2023codebertscore}. To measure the correlation with human preference, we use the CoNaLa dataset~\citep{yin2018mining} and corresponding human annotation on the generated code from various models trained on the dataset~\citep{evtikhiev2022bleu}. To measure the correlation with functional correctness, we use the HumanEval-X dataset~\citep{zheng2023codegeex}. We do not consider \textbf{distinguishability} as an evaluation option, as prior work~\citep{zhou2023codebertscore} has shown it to be an unreliable meta-metric that cannot substitute for execution-based or human-based ratings.

\subsection{Automatic Evaluation Metric Baselines}
The baseline metrics we include can be classified into two groups: \textbf{string-based} and \textbf{neural-model-based} evaluation.

\paragraph{String-based Evaluation}
Most evaluation metrics in code generation have been adapted from natural language generation (NLG) and rely on comparing the generated code to reference code. The most commonly used metric is BLEU~\citep{papineni2002bleu}, which computes the overlaps of $n$-grams in the generated output with those in the reference, where the $n$-grams are tokenized using a language-specific tokenizer~\citep{post-2018-call}. Other metrics include ROUGE-L~\citep{lin2004rouge}, a recall-oriented metric that looks for the longest common subsequence between the reference and the generated code, and METEOR~\citep{banerjee2005meteor}, which is based on unigram matching between the generated code and the reference. However, studies have shown that BLEU may yield similar results for models with different quality levels from the perspective of human graders in code generation~\citep{evtikhiev2022bleu}, leading to the proposal of new evaluation metrics such as RUBY~\citep{tran2019does}. RUBY takes the code structure into account and compares the program dependency graphs (PDG) of the reference and the candidate. If the PDG is impossible to build, the metric falls back to comparing the abstract syntax tree (AST), and if the AST is also impossible to build, it compares the weighted string edit distance between the tokenized reference and candidate sequence. Another recent metric is CodeBLEU~\citep{ren2020codebleu}, which is a composite metric that computes a weighted average of four sub-metrics treating code differently: as a data-flow graph, as an abstract syntax tree, and as text. CodeBLEU is designed to evaluate the quality of generated code for code generation, code translation, and code refinement tasks.

\paragraph{Neural-model-based Evaluation}
Neural-model-based evaluation is becoming increasingly important for evaluating the quality of code generated by deep learning models. CodeBERTScore~\citep{zhou2023codebertscore} is one of the latest approaches that leverages pre-trained code models like CodeBERT~\citep{feng2020codebert} and best practices from natural language generation evaluation to assess the quality of generated code. CodeBERTScore encodes the generated code and reference code independently and considers the natural language context, contextual information of each token, and implementation diversity. It enables the comparison of code pairs that are lexically different and calculates precision and recall based on the best-matching token vector pairs. This approach provides an effective way to evaluate the effectiveness of deep learning models for code intelligence tasks. Note that the authors of CodeBERTScore provided both F1 and F3 scores, with the optional source input. Therefore, we use these four language-specific variants of CodeBERTScore in our experiments.
\begin{table}[]
    \centering
    \resizebox{\columnwidth}{!}{
    \begin{tabular}{l ccc ccc}\toprule
     \multirow{2}{*}{\textbf{Metric}}
     & \multicolumn{3}{c}{\textbf{Example}}
     & \multicolumn{3}{c}{\textbf{Corpus}}\\
     & $\tau$ & $r_p$ & $r_s$ & $\tau$ & $r_p$ & $r_s$ \\\midrule
     BLEU  & .439 & .522 & .488 & .423 & .572 & .542 \\
     CodeBLEU  & .292 & .363 & .331 & .259 & .397 & .339 \\
     chrF & .458 & .570 & .515 & .449 & \underline{.592} & .578 \\
     ROUGE-L & .447 & .529 & .499 & .432 & .581 & .552 \\
     METEOR & .410 & .507 & .462 & .415 & .557 & .534 \\
     RUBY & .331 & .397 & .371 & .339 & .493 & .439 \\
     CodeBERTScore-F1  & .500 & \underline{.609} & .556 & \underline{.464} & .579 & \underline{.595} \\
     CodeBERTScore-F3  & \underline{.505} & \underline{.609} & \underline{.563} & .437 & .549 & .564 \\\midrule
     \texttt{ICE-Score} & \textbf{.556} & .613 & \textbf{.594} & \textbf{.546} & .649 & \textbf{.635} \\
     \texttt{Ref-ICE-Score} & .554 & \textbf{.617} & .591 & .539 & \textbf{.661} & .630 \\\bottomrule
    \end{tabular}}
    \caption{Example-level and corpus-level Kendall-Tau ($\tau$), Pearson ($r_p$) and Spearman ($r_s$) correlations with the human preferred usefulness on CoNaLa.  \texttt{ICE-Score}: without reference code inputs, or reference-free; \texttt{Ref-ICE-Score}: reference-enhanced. The best performance is \textbf{bold}. The second-best performance is \underline{underlined}.}
    \label{tab:conala}
\end{table}

\begin{table*}[!t]
\centering
\resizebox{0.7\textwidth}{!}{
\begin{tabular}{l cc cc cc cc|cc}\toprule
\multirow{2}{*}{\textbf{Metric}}
& \multicolumn{2}{c}{\textbf{Java}}
& \multicolumn{2}{c}{\textbf{C++}}
& \multicolumn{2}{c}{\textbf{Python}}
& \multicolumn{2}{c}{\textbf{JavaScript}}
& \multicolumn{2}{c}{\textbf{Average}}
\\
& $\tau$  & $r_s$ 
& $\tau$  & $r_s$
& $\tau$  & $r_s$ 
& $\tau$  & $r_s$ 
& $\tau$  & $r_s$
\\\midrule

BLEU
& .337 & .401
& .146 & .174
& .251 & .297
& .168 & .199
& .225 & .268
\\
CodeBLEU
& .355 & .421
& .157 & .187
& .272 & .323
& .226 & .267
& \underline{.253} & \underline{.299}
\\
chrF
& .346 & .413
& .166 & .198
& .262 & .312
& .186 & .220
& .240 & .286
\\
ROUGE-L
& .327 & .389
& .143 & .171
& .240 & .284
& .151 & .179
& .215 & .256
\\
METEOR
& .358 & .425
& \underline{.174} & \underline{.208}
& \underline{.276} & \underline{.327}
& \underline{.195} & \underline{.231}
& .251 & .298
\\
RUBY
& .340 & .401
& .139 & .165
& .216 & .255
& .138 & .163
& .208 & .246
\\
CodeBERTScore-F1 
& .314 & .375
& .148 & .177
& .231 & .276
& .145 & .172
& .209 & .250
\\
CodeBERTScore-F3 
& .356 & .426
& .166 & .198
& .262 & .312
& .189 & .226
& .243 & .291
\\\midrule
\texttt{ICE-Score}
& \textbf{.427} & \textbf{.442}
& \textbf{.320} & \textbf{.326}
& .279 & .282
& .316 & .321
& \textbf{.336} & \textbf{.343}
\\
\texttt{Ref-ICE-Score}
& .388 & .404
& .274 & .282
& \textbf{.318} & \textbf{.325}
& \textbf{.340} & \textbf{.348}
& .330 & .340
\\

\bottomrule
\end{tabular}}
\caption{Example-level Kendall-Tau ($\tau$) and Spearman ($r_s$) correlations with the execution-based functional
correctness on HumanEval.  \texttt{ICE-Score}: without reference code inputs, or reference-free; \texttt{Ref-ICE-Score}: with reference code inputs, or reference-enhanced. The best performance is \textbf{bold}. The second-best performance is \underline{underlined}.}
\label{tab:humaneval_example}
\end{table*}

\subsection{Datasets and Evaluation Aspects}
\paragraph{Human-based Usefulness Experiments} Similar to \cite{zhou2023codebertscore}, we conduct an evaluation on the CoNaLa benchmark~\citep{yin2018mining}, which is a widely used dataset for natural language context to Python code generation. To measure the correlation between each evaluation metric and human preference, we utilize the human annotations provided by \cite{evtikhiev2022bleu}. Specifically, for each example in the dataset, experienced software developers were asked to grade the generated code snippets from five different models. The grading scale ranges from zero to four, with zero indicating that the generated code is irrelevant and unhelpful, and four indicating that the generated code solves the problem accurately. The dataset comprises a total of 2,860 annotated code snippets (5 generations $\times$ 472 examples) with each snippet being graded by 4.5 annotators on average.

\paragraph{Execution-based Functional Correctness Experiments} We conduct an evaluation of functional correctness using the HumanEval benchmark~\citep{chen2021evaluating}, which provides natural language goals, input-output test cases, and reference solutions written by humans for each example. The benchmark originally consists of 164 coding problems in Python, and has been extended by \cite{cassano:multipl-e} to 18 other programming languages, including Java, C++, Python, and JavaScript. We chose to evaluate our models on these languages, as they are among the most popular programming languages. The translated examples also include the predictions of \texttt{code-davinci-002} and their corresponding functional correctness scores. Inspired by \cite{zhou2023codebertscore}, we obtain them from the HumanEval-X dataset~\citep{zheng2023codegeex}. As each problem has nearly 200 generated code samples on average, it would be computationally expensive to evaluate them all using LLMs. Therefore, we randomly select 20 samples from each problem, and collect all samples from problems where no more than 20 versions of code were generated.

\paragraph{Correlation Metrics} To measure the correlation between each metric's scores and the references, we follow best practices in natural language evaluation and used Kendall-Tau ($\tau$), Pearson ($r_p$), and Spearman ($r_s$) coefficients.\footnote{We use the implementations from \url{https://scipy.org/}}. To systematically study the efficacy of each automatic evaluation metric, we compute both example-level and corpus-level correlations. The example-level correlation is the average correlation of each problem example, while the corpus-level correlation is the correlation of all aggregated examples in the task.

\begin{table*}[]
\centering
\resizebox{0.7\textwidth}{!}{
\begin{tabular}{l cc cc cc cc|cc}\toprule
\multirow{2}{*}{\textbf{Metric}}
& \multicolumn{2}{c}{\textbf{Java}}
& \multicolumn{2}{c}{\textbf{C++}}
& \multicolumn{2}{c}{\textbf{Python}}
& \multicolumn{2}{c}{\textbf{JavaScript}}
& \multicolumn{2}{c}{\textbf{Average}}
\\
& $\tau$  & $r_s$ 
& $\tau$  & $r_s$
& $\tau$  & $r_s$ 
& $\tau$  & $r_s$ 
& $\tau$  & $r_s$
\\\midrule

BLEU
& .267 & .326
& .225 & .276
& .281 & .344
& .220 & .270
& .248 & .304
\\
CodeBLEU
& .293 & .359
& .212 & .260
& .303 & .371
& \underline{.315} & \underline{.385}
& .281 & .343
\\
chrF
& .290 & .355
& .266 & .325
& .328 & .402
& .279 & .342
& .291 & .356
\\
ROUGE-L
& .280 & .342
& .234 & .286
& .296 & .363
& .216 & .264
& .256 & .314
\\
METEOR
& .318 & .389
& .260 & .319
& .349 & .427
& .311 & .380
& .309 & .379
\\
RUBY
& .276 & .337
& .219 & .268
& .279 & .341
& .219 & .268
& .248 & .303
\\
CodeBERTScore-F1 
& .244 & .298
& .219 & .268
& .264 & .324
& .214 & .262
& .235 & .288
\\
CodeBERTScore-F3 
& .281 & .344
& .243 & .297
& .313 & .384
& .261 & .320
& .275 & .336
\\\midrule
\texttt{ICE-Score}
& .330 & .345
& .313 & .321
& .294 & .298
& .315 & .323
& .313 & .322
\\
\texttt{Ref-ICE-Score}
& \textbf{.412} & \textbf{.438}
& \textbf{.367} & \textbf{.383}
& \textbf{.425} & \textbf{.446}
& \textbf{.432} & \textbf{.455}
& \textbf{.409} & \textbf{.431}
\\

\bottomrule
\end{tabular}}
\caption{Corpus-level Kendall-Tau ($\tau$) and Spearman ($r_s$) correlations with the execution-based functional
correctness on HumanEval.  \texttt{ICE-Score}: without reference code inputs, or reference-free; \texttt{Ref-ICE-Score}: with reference code inputs, or reference-enhanced. The best performance is \textbf{bold}. The second-best performance is \underline{underlined}.}
\label{tab:humaneval_corpus}
\end{table*}
\section{Results}
\paragraph{Human-based Usefulness} Table~\ref{tab:conala}  shows the correlation between different metrics with human preference. We compare two variants of our evaluation approach, reference-free and reference-enhanced evaluations, with 10 baseline metrics and their variants. We find that \texttt{ICE-Score} outperform these metrics by a significant margin, regarding both example- and corpus-level correlations. Our observation is consistent with the work of CodeBERScore, where the variants of CodeBERScore mostly outperform the strong baselines like chrF and ROUGE-L. For example, \texttt{ICE-Score} achieves 0.556 and 0.546 measured by Spearman correlation on example level and corpus level, respectively. In contrast, prior evaluation metrics barely reach a score of 0.5. In addition, we find that \texttt{Ref-ICE-Score} does not significantly improve the performance, indicating the reference code may not be optimized. Our further analysis of the human rating of CoNaLa reference code complies with this implication, where the average score of the reference code only achieves 3.4 out of 4, suggesting that not all human practitioners consider the reference fully useful.

\paragraph{Execution-based Functional Correctness} Table~\ref{tab:humaneval_example} and Table~\ref{tab:humaneval_corpus} present the results of example- and corpus-level functional correctness, respectively. From Table~\ref{tab:humaneval_example}, we observe that both reference-free and reference-enhanced \texttt{Ref-ICE-Scores}consistently outperform the other baselines across all four programming languages on the example level. \texttt{ICE-Score} even outperforms the reference-enhanced one, suggesting potential bias in some reference code. Additionally, we find that METEOR and CodeBLEU receive better correlations than all variants of CodeBERTScore, indicating that they are still strong baselines compared to the recent neural-model-based evaluators in code generation. In Table~\ref{tab:humaneval_corpus}, we observe that our \texttt{Ref-ICE-Score} achieves state-of-the-art performance among all evaluation metrics. When compared to other baselines, \texttt{ICE-Score} still achieves comparable results to the source-free CodeBERTScore-F3.

\begin{figure*}[!t]
    \centering
    \resizebox{0.8\textwidth}{!}{\includegraphics{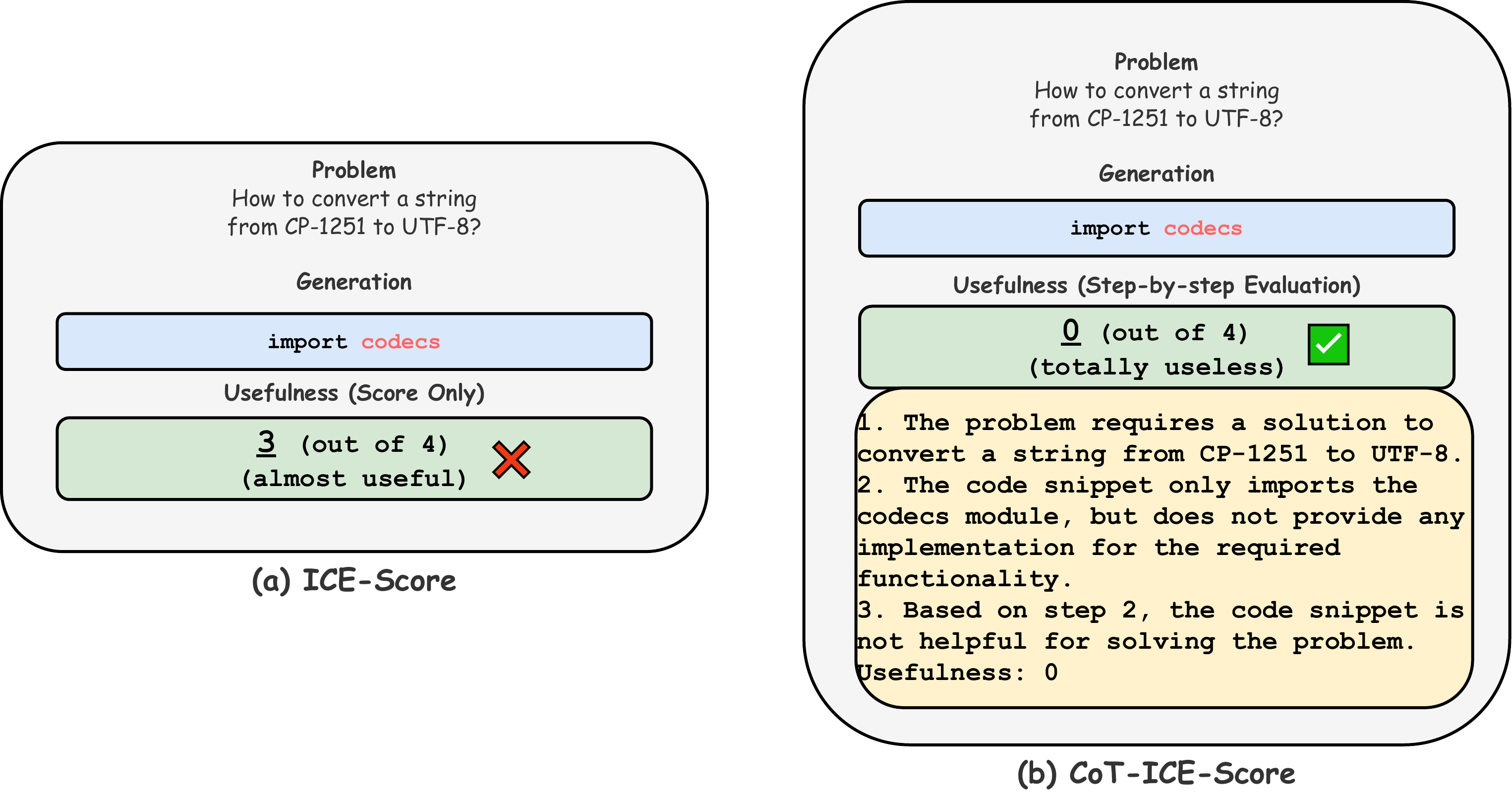}}
    \caption{Example inputs and outputs with (a) \texttt{ICE-Score}, (b) \texttt{ICE-Score} with Zero-Shot Chain-of-Thought. With the step-by-step evaluation, the output assessment is more aligned with human preference.}
    \label{fig:cot_conala}
\end{figure*}

\section{Ablation Study}

\begin{table}[]
\centering
\resizebox{\columnwidth}{!}{
\begin{tabular}{l ccc ccc}\toprule
 \multirow{2}{*}{\textbf{Metric}}
 & \multicolumn{3}{c}{\textbf{Example}}
 & \multicolumn{3}{c}{\textbf{Corpus}}\\
 & $\tau$ & $r_p$ & $r_s$ & $\tau$ & $r_p$ & $r_s$ \\\midrule
{\texttt{ICE-Score}} & .556 & .613 & .594 & .546 & .649 & .635 \\
\rowcolor{shadecolor} \texttt{CoT-ICE-Score} & \textbf{.561} & \textbf{.628} & \textbf{.600} & \textbf{.579} & \textbf{.703} & \textbf{.665}\\
\texttt{Ref-ICE-Score} & .554 & .617 & .591 & .539 & .661 & .630\\
\rowcolor{shadecolor} \texttt{CoT-Ref-ICE-Score}& \textbf{.571} &\textbf{.639} &\textbf{.607} &\textbf{.583} &\textbf{.712} &\textbf{.667}\\\bottomrule
\end{tabular}}
\caption{Example-level and corpus-level Kendall-Tau ($\tau$), Pearson ($r_p$) and Spearman ($r_s$) correlations with the human preferred usefulness on CoNaLa. \texttt{ICE-Score}: without reference code inputs, or reference-free; \texttt{Ref-ICE-Score}: with reference code inputs, or reference-enhanced. \texttt{CoT-} indicates the use of ZS-CoT. The best performance is \textbf{bold}.}
\label{tab:cot_conala}
\end{table}
\paragraph{Does reasoning help the code evaluation?} Prior work~\citep{weichain,kojimalarge} has demonstrated that the performance of LLMs can be significantly improved via Chain-of-Thought (CoT) and Zero-Shot-Chain-of-Thought (ZS-CoT), where the prompts instruct LLMs to perform the task in a step-by-step manner. Here, we explore the zero-shot reasoning ability of LLMs in evaluating code generation. Specifically, we instruct GPT-3.5 to perform CoT-evaluation by adding "Step-by-step Evaluation:" at the end of the prompt. An example of the zero-shot-CoT prompt is shown in Figure~\ref{fig:cot_conala}. Instead of using LLMs to extract the evaluation score from the reasoning steps, like the original metric of zero-shot-CoT via multiple queries, we design a rule-based parser to extract scores. Due to limited resources, we only evaluate on CoNaLa in Table~\ref{tab:cot_conala}. Our results show that ZS-CoT can significantly improve the reliability of code evaluation. Additionally, we find that \texttt{Ref-ICE-Score} can achieve better results than reference-free ones via ZS-CoT, even though their performances are similar without CoT processing. This suggests that LLMs can exploit the use of reference code through reasoning.

\begin{table}[!h]
\centering
\resizebox{\columnwidth}{!}{

\begin{tabular}{l ccc ccc}\toprule
 \multirow{2}{*}{\textbf{Metric}}
 & \multicolumn{3}{c}{\textbf{Example}}
 & \multicolumn{3}{c}{\textbf{Corpus}}\\
 & $\tau$ & $r_p$ & $r_s$ & $\tau$ & $r_p$ & $r_s$ \\\midrule
{\texttt{ICE-Score-3.5}} & .556 & .613 & .594 & .546 & .649 & .635 \\
\rowcolor{shadecolor} \texttt{ICE-Score-4} & \textbf{.612} & \textbf{.658} & \textbf{.611} & \textbf{.592} & \textbf{.720}& \textbf{.688}\\
\texttt{Ref-ICE-Score-3.5} & .554 & .617 & .591 & .539 & .661 & .630\\
\rowcolor{shadecolor} \texttt{Ref-ICE-Score-4}& \textbf{.592} &\textbf{.647} &\textbf{.634} &\textbf{.632} &\textbf{.744} &\textbf{.690}\\\bottomrule
\end{tabular}}
\caption{Example-level and corpus-level Kendall-Tau ($\tau$), Pearson ($r_p$) and Spearman ($r_s$) correlations with the human preferred usefulness on CoNaLa. \texttt{ICE-Score}: without reference code inputs, or reference-free; \texttt{Ref-ICE-Score}: with reference code inputs, or reference-enhanced. \texttt{-3.5} and \texttt{-4} suggest the different backbone models. The best performance is \textbf{bold}.}
\label{tab:gpt4_conala}
\end{table}

\paragraph{Does more-capable backbone LLM yield better performance on code evaluation?} As shown in previous studies~\cite{openai2023gpt4,bubeck2023sparks}, GPT-4 significantly outperforms GPT-3.5 on various tasks. Therefore, we use GPT-4 as the backbone model for \texttt{ICE-Score} and evaluate its performance on CoNaLa. The results in Table~\ref{tab:gpt4_conala} indicate that GPT-4 consistently surpasses \texttt{GPT-3.5-turbo} on evaluating code, suggesting it has the superior capability of code comprehension. We also note that using a more capable model like GPT-4 can guarantee even better performance, compared to using ZS-CoT techniques in Table~\ref{tab:cot_conala}.
\section{Discussion}
\label{sec:discussion}

\begin{table*}[!th]
    \centering
    \resizebox{0.8\textwidth}{!}{
    \begin{tabular}{lcc}\toprule
    \textbf{Dataset} & \textbf{Release Year} & \textbf{Likely to be contaminated?}\\\midrule
      CoNaLa   & 2018 & \cmark\\
      human-annotated CoNaLa  w/ generated code  & 2023 & \xmark \\
      HumanEval (Python) & 2021 & \cmark\\
      HumanEval-X (w/o Python) & 2023 &\xmark\\
      human-annotated HumanEval-X w/ generated code  & 2023 & \xmark\\\bottomrule
    \end{tabular}}
    \caption{Dataset, Release Year and the likelihood of data contamination for each dataset used in our study.}
    \label{tab:data_contamination}
\end{table*}

\paragraph{Data Contamination} Evaluations on recent closed-source LLMs have been criticized for the possibility of data contamination~\citep{aiyappa2023can}, where the model may have already seen the evaluation datasets during training, due to the opaque training details of these models. For instance, \citet{kocmi2023large} conducted an empirical study on a few closed-source LLMs, including GPT-3.5, and suggested that LLMs are the state-of-the-art evaluators of translation quality, based on the evaluation of the WMT22 Metric Shared Task~\citep{freitag2022results}. However, as most of the evaluated models were trained on data prior to 2022\footnote{\url{https://platform.openai.com/docs/model-index-for-researchers}}, it is highly likely that these models have been trained with some human-rated translation quality data. Similarly, G-EVAL\citep{liu2023gpteval} shows that GPT-3.5 and GPT-4 are the state-of-the-art evaluators of natural language generation (NLG) with the evaluation of three NLG datasets. However, as these human-annotated datasets were released before 2021, it is probable that they were included in the training data of GPT-3.5 and GPT-4. In contrast, our work is minimally impacted by data contamination, as we report the data release year in Table~\ref{tab:data_contamination}. Our analysis suggests that only CoNaL and HumanEval (Python) datasets may have been contaminated, and it is unlikely that GPT-3.5 has seen any human annotation or generated code during training. 

\paragraph{Human-aligned Evaluation Beyond Code Generation} While our study has shown that LLMs can achieve state-of-the-art performance in evaluating the functional correctness and usefulness of generated source code, the question remains as to whether LLMs can be utilized to evaluate code intelligence tasks beyond code generation. \citet{allamanis2018survey} have identified several downstream applications such as code translation, commit message generation, and code summarization. While some studies have investigated the human evaluation of these tasks, none of them have released the annotation data or fully described the human evaluation criteria. This presents a challenge for analyzing if \texttt{ICE-Score} can be adapted to these tasks. For example, \citet{hu2022correlating} proposed a human evaluation metric for code documentation generation quality, which is specifically designed for code comment generation and commit message generation. Their metric includes three aspects: \textit{Language-related}, \textit{Content-related}, and \textit{Effectiveness-related}, with detailed task descriptions and explanations of assigned scores. We propose that the information provided in their metric can be used to create prompts for LLM-based evaluation and enable human-aligned evaluation of code documentation generation.

\section{Related Work}

\paragraph{Large Language Models for Code.} LLMs pre-trained on large-scale code data have demonstrated strong capabilities in code intelligence tasks, such as code completion~\cite{li2023starcoder,luo2023wizardcoder,roziere2023code}, code summarization~\cite{ahmed2022few,sun2023automatic} and program repair~\cite{surameery2023use,sobania2023analysis}. However, they remain unreliable, particularly in scenarios that require an understanding of natural language. Recent studies~\cite{muennighoff2023scaling,matraining} show that pre-training on both text and code results in the optimal model performance on natural language and code understanding. Furthermore, in order to make LLMs more human-aligned and more capable of performing complex tasks, instruction tuning is proposed to enhance the capability of following natural language requirements. In this work, we utilize such instruction-tuned LLMs to conduct multi-dimensional code evaluation via various instructions.

\paragraph{Automatic Evaluation Metrics for Generation.} The quest for reliable and robust automatic evaluation metrics for generated content has been a cornerstone in natural language processing. Traditionally, string-based metrics such as BLEU~\cite{papineni2002bleu}, ROUGE~\cite{lin2004rouge}, and METEOR~\cite{banerjee2005meteor} have dominated the landscape, primarily when assessing machine translation or text summarization outputs. While these metrics provide a quick and cost-effective means of evaluating the quality of the generated text, they often fall short of capturing the nuanced intricacies and semantic richness inherent in natural language. To mitigate such drawbacks, a few neural-based multi-dimensional evaluation metrics have been proposed for text generation, such as UniEval~\cite{zhong2022towards}, GPTScore~\cite{fu2023gptscore} and G-EVAL~\cite{liu2023gpteval}. However, when it comes to code generation, where both syntactical correctness and semantic intent are paramount, there are few attempts to address these challenges. Instead, the most dominant metrics still compute the similarity between generated code and reference code.  In this work, we introduce \texttt{ICE-Score}, a novel metric that not only addresses the limitations of its predecessors but also harnesses the capabilities of LLMs, setting a new benchmark for the evaluation of code generation tasks.
\section{Conclusion}

In this paper, we propose a novel evaluation metric based on large language models trained on both text and code, which can better capture the complex syntax and semantics of code intelligence tasks. Our metric achieves superior correlations with subjective human preferences and objective functional correctness, both at the example and corpus levels, without reference and test suites. We conduct an extensive evaluation of four programming languages (Java, Python, C, C++, and JavaScript) and demonstrate the effectiveness of our proposed method on human-based usefulness and execution-based functional correctness. We have publicly released our evaluation metrics and datasets to encourage the development of more accurate and effective evaluation metrics for tasks involving source code.

\section*{Acknowledgements}
We thank Haolan Zhan and Yufei Wang for the helpful feedback on the paper.

\section*{Limitations}
Our proposed evaluation metric is based on the assumption that LLMs can follow the instructions to evaluate the code snippets. The backbone models we investigated are closed-source state-of-the-art LLMs from OepnAI. As we noticed that there is a huge performance gap between current closed-source and open-source LLMs, it is possible that \texttt{ICE-Score} can be adapted with an open-source LLM trained on code and text, such as WizardCoder~\cite{luo2023wizardcoder} and OctoPack~\cite{muennighoff2023octopack}. Hence, we encourage future investigations on open-source LLMs for code evaluation. In addition, as discussed in Section~\ref{sec:discussion}, our experiments only focus on two code generation tasks. There are other code intelligence tasks like program repair and code summarization. However, due to the limited study on human evaluation of these tasks, no open-source dataset is publicly available or documented in detail. Finally, \texttt{ICE-Score} assumes that either model weights or model APIs are available, which is costly for some users. We, therefore, suggest future work on proposing low-cost evaluation metrics.
\bibliography{custom}

\begin{thebibliography}{47}
\expandafter\ifx\csname natexlab\endcsname\relax\def\natexlab#1{#1}\fi

\bibitem[{Ahmed and Devanbu(2022)}]{ahmed2022few}
Toufique Ahmed and Premkumar Devanbu. 2022.
\newblock Few-shot training llms for project-specific code-summarization.
\newblock In \emph{Proceedings of the 37th IEEE/ACM International Conference on Automated Software Engineering}, pages 1--5.

\bibitem[{Aiyappa et~al.(2023)Aiyappa, An, Kwak, and Ahn}]{aiyappa2023can}
Rachith Aiyappa, Jisun An, Haewoon Kwak, and Yong-Yeol Ahn. 2023.
\newblock Can we trust the evaluation on chatgpt?
\newblock \emph{arXiv preprint arXiv:2303.12767}.

\bibitem[{Allal et~al.(2023)Allal, Li, Kocetkov, Mou, Akiki, Ferrandis, Muennighoff, Mishra, Gu, Dey et~al.}]{allal2023santacoder}
Loubna~Ben Allal, Raymond Li, Denis Kocetkov, Chenghao Mou, Christopher Akiki, Carlos~Munoz Ferrandis, Niklas Muennighoff, Mayank Mishra, Alex Gu, Manan Dey, et~al. 2023.
\newblock Santacoder: don't reach for the stars!
\newblock \emph{arXiv preprint arXiv:2301.03988}.

\bibitem[{Allamanis et~al.(2018)Allamanis, Barr, Devanbu, and Sutton}]{allamanis2018survey}
Miltiadis Allamanis, Earl~T Barr, Premkumar Devanbu, and Charles Sutton. 2018.
\newblock A survey of machine learning for big code and naturalness.
\newblock \emph{ACM Computing Surveys (CSUR)}, 51(4):1--37.

\bibitem[{Banerjee and Lavie(2005)}]{banerjee2005meteor}
Satanjeev Banerjee and Alon Lavie. 2005.
\newblock Meteor: An automatic metric for mt evaluation with improved correlation with human judgments.
\newblock In \emph{Proceedings of the acl workshop on intrinsic and extrinsic evaluation measures for machine translation and/or summarization}, pages 65--72.

\bibitem[{Brown et~al.(2020)Brown, Mann, Ryder, Subbiah, Kaplan, Dhariwal, Neelakantan, Shyam, Sastry, Askell et~al.}]{brown2020language}
Tom Brown, Benjamin Mann, Nick Ryder, Melanie Subbiah, Jared~D Kaplan, Prafulla Dhariwal, Arvind Neelakantan, Pranav Shyam, Girish Sastry, Amanda Askell, et~al. 2020.
\newblock Language models are few-shot learners.
\newblock \emph{Advances in neural information processing systems}, 33:1877--1901.

\bibitem[{Bubeck et~al.(2023)Bubeck, Chandrasekaran, Eldan, Gehrke, Horvitz, Kamar, Lee, Lee, Li, Lundberg et~al.}]{bubeck2023sparks}
S{\'e}bastien Bubeck, Varun Chandrasekaran, Ronen Eldan, Johannes Gehrke, Eric Horvitz, Ece Kamar, Peter Lee, Yin~Tat Lee, Yuanzhi Li, Scott Lundberg, et~al. 2023.
\newblock Sparks of artificial general intelligence: Early experiments with gpt-4.
\newblock \emph{arXiv preprint arXiv:2303.12712}.

\bibitem[{Cassano et~al.(2023)Cassano, Gouwar, Nguyen, Nguyen, Phipps-Costin, Pinckney, Yee, Zi, Anderson, Feldman, Guha, Greenberg, and Jangda}]{cassano:multipl-e}
Federico Cassano, John Gouwar, Daniel Nguyen, Sydney Nguyen, Luna Phipps-Costin, Donald Pinckney, Ming-Ho Yee, Yangtian Zi, Carolyn~Jane Anderson, Molly~Q Feldman, Arjun Guha, Michael Greenberg, and Abhinav Jangda. 2023.
\newblock {MultiPL-E}: A scalable and polyglot approach to benchmarking neural code generation.
\newblock \emph{{IEEE} Transactions of Software Engineering (TSE)}.

\bibitem[{Chen et~al.(2021)Chen, Tworek, Jun, Yuan, Pinto, Kaplan, Edwards, Burda, Joseph, Brockman et~al.}]{chen2021evaluating}
Mark Chen, Jerry Tworek, Heewoo Jun, Qiming Yuan, Henrique Ponde de~Oliveira Pinto, Jared Kaplan, Harri Edwards, Yuri Burda, Nicholas Joseph, Greg Brockman, et~al. 2021.
\newblock Evaluating large language models trained on code.
\newblock \emph{arXiv preprint arXiv:2107.03374}.

\bibitem[{Chen et~al.(2023)Chen, Lin, Sch{\"a}rli, and Zhou}]{chen2023teaching}
Xinyun Chen, Maxwell Lin, Nathanael Sch{\"a}rli, and Denny Zhou. 2023.
\newblock Teaching large language models to self-debug.
\newblock \emph{arXiv preprint arXiv:2304.05128}.

\bibitem[{Eghbali and Pradel(2022)}]{eghbali2022crystalbleu}
Aryaz Eghbali and Michael Pradel. 2022.
\newblock Crystalbleu: precisely and efficiently measuring the similarity of code.
\newblock In \emph{37th IEEE/ACM International Conference on Automated Software Engineering}, pages 1--12.

\bibitem[{Evtikhiev et~al.(2023)Evtikhiev, Bogomolov, Sokolov, and Bryksin}]{evtikhiev2022bleu}
Mikhail Evtikhiev, Egor Bogomolov, Yaroslav Sokolov, and Timofey Bryksin. 2023.
\newblock Out of the bleu: how should we assess quality of the code generation models?
\newblock \emph{Journal of Systems and Software}, 203:111741.

\bibitem[{Feng et~al.(2020)Feng, Guo, Tang, Duan, Feng, Gong, Shou, Qin, Liu, Jiang et~al.}]{feng2020codebert}
Zhangyin Feng, Daya Guo, Duyu Tang, Nan Duan, Xiaocheng Feng, Ming Gong, Linjun Shou, Bing Qin, Ting Liu, Daxin Jiang, et~al. 2020.
\newblock Codebert: A pre-trained model for programming and natural languages.
\newblock In \emph{Findings of the Association for Computational Linguistics: EMNLP 2020}, pages 1536--1547.

\bibitem[{Freitag et~al.(2022)Freitag, Rei, Mathur, Lo, Stewart, Avramidis, Kocmi, Foster, Lavie, and Martins}]{freitag2022results}
Markus Freitag, Ricardo Rei, Nitika Mathur, Chi-kiu Lo, Craig Stewart, Eleftherios Avramidis, Tom Kocmi, George Foster, Alon Lavie, and Andr{\'e}~FT Martins. 2022.
\newblock Results of wmt22 metrics shared task: Stop using bleu--neural metrics are better and more robust.
\newblock In \emph{Proceedings of the Seventh Conference on Machine Translation (WMT)}, pages 46--68.

\bibitem[{Fu et~al.(2023)Fu, Ng, Jiang, and Liu}]{fu2023gptscore}
Jinlan Fu, See-Kiong Ng, Zhengbao Jiang, and Pengfei Liu. 2023.
\newblock Gptscore: Evaluate as you desire.
\newblock \emph{arXiv preprint arXiv:2302.04166}.

\bibitem[{Hindle et~al.(2016)Hindle, Barr, Gabel, Su, and Devanbu}]{hindle2016naturalness}
Abram Hindle, Earl~T Barr, Mark Gabel, Zhendong Su, and Premkumar Devanbu. 2016.
\newblock On the naturalness of software.
\newblock \emph{Communications of the ACM}, 59(5):122--131.

\bibitem[{Hu et~al.(2022)Hu, Chen, Wang, Xia, Lo, and Zimmermann}]{hu2022correlating}
Xing Hu, Qiuyuan Chen, Haoye Wang, Xin Xia, David Lo, and Thomas Zimmermann. 2022.
\newblock Correlating automated and human evaluation of code documentation generation quality.
\newblock \emph{ACM Transactions on Software Engineering and Methodology (TOSEM)}, 31(4):1--28.

\bibitem[{Huang et~al.(2022)Huang, Gu, Hou, Wu, Wang, Yu, and Han}]{huang2022large}
Jiaxin Huang, Shixiang~Shane Gu, Le~Hou, Yuexin Wu, Xuezhi Wang, Hongkun Yu, and Jiawei Han. 2022.
\newblock Large language models can self-improve.
\newblock \emph{arXiv preprint arXiv:2210.11610}.

\bibitem[{Kocmi and Federmann(2023)}]{kocmi2023large}
Tom Kocmi and Christian Federmann. 2023.
\newblock \href {http://arxiv.org/abs/2302.14520} {Large language models are state-of-the-art evaluators of translation quality}.

\bibitem[{Kocmi et~al.(2021)Kocmi, Federmann, Grundkiewicz, Junczys-Dowmunt, Matsushita, and Menezes}]{kocmi-etal-2021-ship}
Tom Kocmi, Christian Federmann, Roman Grundkiewicz, Marcin Junczys-Dowmunt, Hitokazu Matsushita, and Arul Menezes. 2021.
\newblock \href {https://aclanthology.org/2021.wmt-1.57} {To ship or not to ship: An extensive evaluation of automatic metrics for machine translation}.
\newblock In \emph{Proceedings of the Sixth Conference on Machine Translation}, pages 478--494, Online. Association for Computational Linguistics.

\bibitem[{Kojima et~al.()Kojima, Gu, Reid, Matsuo, and Iwasawa}]{kojimalarge}
Takeshi Kojima, Shixiang~Shane Gu, Machel Reid, Yutaka Matsuo, and Yusuke Iwasawa.
\newblock Large language models are zero-shot reasoners.
\newblock In \emph{Advances in Neural Information Processing Systems}.

\bibitem[{Li et~al.(2023)Li, Allal, Zi, Muennighoff, Kocetkov, Mou, Marone, Akiki, Li, Chim et~al.}]{li2023starcoder}
Raymond Li, Loubna~Ben Allal, Yangtian Zi, Niklas Muennighoff, Denis Kocetkov, Chenghao Mou, Marc Marone, Christopher Akiki, Jia Li, Jenny Chim, et~al. 2023.
\newblock Starcoder: may the source be with you!
\newblock \emph{arXiv preprint arXiv:2305.06161}.

\bibitem[{Lin(2004)}]{lin2004rouge}
Chin-Yew Lin. 2004.
\newblock Rouge: A package for automatic evaluation of summaries.
\newblock In \emph{Text summarization branches out}, pages 74--81.

\bibitem[{Liu et~al.(2023)Liu, Iter, Xu, Wang, Xu, and Zhu}]{liu2023gpteval}
Yang Liu, Dan Iter, Yichong Xu, Shuohang Wang, Ruochen Xu, and Chenguang Zhu. 2023.
\newblock Gpteval: Nlg evaluation using gpt-4 with better human alignment.
\newblock \emph{arXiv preprint arXiv:2303.16634}.

\bibitem[{Luo et~al.(2023)Luo, Xu, Zhao, Sun, Geng, Hu, Tao, Ma, Lin, and Jiang}]{luo2023wizardcoder}
Ziyang Luo, Can Xu, Pu~Zhao, Qingfeng Sun, Xiubo Geng, Wenxiang Hu, Chongyang Tao, Jing Ma, Qingwei Lin, and Daxin Jiang. 2023.
\newblock Wizardcoder: Empowering code large language models with evol-instruct.
\newblock \emph{arXiv preprint arXiv:2306.08568}.

\bibitem[{Ma et~al.()Ma, Liu, Yu, Zhang, Jiang, Wang, and Li}]{matraining}
Yingwei Ma, Yue Liu, Yue Yu, Yuanliang Zhang, Yu~Jiang, Changjian Wang, and Shanshan Li.
\newblock At which training stage does code data help llms reasoning?

\bibitem[{Madaan et~al.(2023)Madaan, Tandon, Gupta, Hallinan, Gao, Wiegreffe, Alon, Dziri, Prabhumoye, Yang et~al.}]{madaan2023self}
Aman Madaan, Niket Tandon, Prakhar Gupta, Skyler Hallinan, Luyu Gao, Sarah Wiegreffe, Uri Alon, Nouha Dziri, Shrimai Prabhumoye, Yiming Yang, et~al. 2023.
\newblock Self-refine: Iterative refinement with self-feedback.
\newblock \emph{arXiv preprint arXiv:2303.17651}.

\bibitem[{Muennighoff et~al.(2023{\natexlab{a}})Muennighoff, Liu, Zebaze, Zheng, Hui, Zhuo, Singh, Tang, von Werra, and Longpre}]{muennighoff2023octopack}
Niklas Muennighoff, Qian Liu, Armel Zebaze, Qinkai Zheng, Binyuan Hui, Terry~Yue Zhuo, Swayam Singh, Xiangru Tang, Leandro von Werra, and Shayne Longpre. 2023{\natexlab{a}}.
\newblock Octopack: Instruction tuning code large language models.
\newblock \emph{arXiv preprint arXiv:2308.07124}.

\bibitem[{Muennighoff et~al.(2023{\natexlab{b}})Muennighoff, Rush, Barak, Scao, Piktus, Tazi, Pyysalo, Wolf, and Raffel}]{muennighoff2023scaling}
Niklas Muennighoff, Alexander~M Rush, Boaz Barak, Teven~Le Scao, Aleksandra Piktus, Nouamane Tazi, Sampo Pyysalo, Thomas Wolf, and Colin Raffel. 2023{\natexlab{b}}.
\newblock Scaling data-constrained language models.
\newblock \emph{arXiv preprint arXiv:2305.16264}.

\bibitem[{OpenAI(2023)}]{openai2023gpt4}
OpenAI. 2023.
\newblock \href {http://arxiv.org/abs/2303.08774} {Gpt-4 technical report}.

\bibitem[{Ouyang et~al.(2022)Ouyang, Wu, Jiang, Almeida, Wainwright, Mishkin, Zhang, Agarwal, Slama, Ray et~al.}]{ouyang2022training}
Long Ouyang, Jeffrey Wu, Xu~Jiang, Diogo Almeida, Carroll Wainwright, Pamela Mishkin, Chong Zhang, Sandhini Agarwal, Katarina Slama, Alex Ray, et~al. 2022.
\newblock Training language models to follow instructions with human feedback.
\newblock \emph{Advances in Neural Information Processing Systems}, 35:27730--27744.

\bibitem[{Papineni et~al.(2002)Papineni, Roukos, Ward, and Zhu}]{papineni2002bleu}
Kishore Papineni, Salim Roukos, Todd Ward, and Wei-Jing Zhu. 2002.
\newblock Bleu: a method for automatic evaluation of machine translation.
\newblock In \emph{Proceedings of the 40th annual meeting of the Association for Computational Linguistics}, pages 311--318.

\bibitem[{Popovi{\'c}(2015)}]{popovic2015chrf}
Maja Popovi{\'c}. 2015.
\newblock chrf: character n-gram f-score for automatic mt evaluation.
\newblock In \emph{Proceedings of the tenth workshop on statistical machine translation}, pages 392--395.

\bibitem[{Post(2018)}]{post-2018-call}
Matt Post. 2018.
\newblock \href {https://doi.org/10.18653/v1/W18-6319} {A call for clarity in reporting {BLEU} scores}.
\newblock In \emph{Proceedings of the Third Conference on Machine Translation: Research Papers}, pages 186--191, Brussels, Belgium. Association for Computational Linguistics.

\bibitem[{Ren et~al.(2020)Ren, Guo, Lu, Zhou, Liu, Tang, Sundaresan, Zhou, Blanco, and Ma}]{ren2020codebleu}
Shuo Ren, Daya Guo, Shuai Lu, Long Zhou, Shujie Liu, Duyu Tang, Neel Sundaresan, Ming Zhou, Ambrosio Blanco, and Shuai Ma. 2020.
\newblock Codebleu: a method for automatic evaluation of code synthesis.
\newblock \emph{arXiv preprint arXiv:2009.10297}.

\bibitem[{Rozi{\`e}re et~al.(2023)Rozi{\`e}re, Gehring, Gloeckle, Sootla, Gat, Tan, Adi, Liu, Remez, Rapin et~al.}]{roziere2023code}
Baptiste Rozi{\`e}re, Jonas Gehring, Fabian Gloeckle, Sten Sootla, Itai Gat, Xiaoqing~Ellen Tan, Yossi Adi, Jingyu Liu, Tal Remez, J{\'e}r{\'e}my Rapin, et~al. 2023.
\newblock Code llama: Open foundation models for code.
\newblock \emph{arXiv preprint arXiv:2308.12950}.

\bibitem[{Sobania et~al.(2023)Sobania, Briesch, Hanna, and Petke}]{sobania2023analysis}
Dominik Sobania, Martin Briesch, Carol Hanna, and Justyna Petke. 2023.
\newblock An analysis of the automatic bug fixing performance of chatgpt.
\newblock \emph{arXiv preprint arXiv:2301.08653}.

\bibitem[{Sun et~al.(2023)Sun, Fang, You, Miao, Liu, Li, Deng, Huang, Chen, Zhang et~al.}]{sun2023automatic}
Weisong Sun, Chunrong Fang, Yudu You, Yun Miao, Yi~Liu, Yuekang Li, Gelei Deng, Shenghan Huang, Yuchen Chen, Quanjun Zhang, et~al. 2023.
\newblock Automatic code summarization via chatgpt: How far are we?
\newblock \emph{arXiv preprint arXiv:2305.12865}.

\bibitem[{Surameery and Shakor(2023)}]{surameery2023use}
Nigar M~Shafiq Surameery and Mohammed~Y Shakor. 2023.
\newblock Use chat gpt to solve programming bugs.
\newblock \emph{International Journal of Information Technology \& Computer Engineering (IJITC) ISSN: 2455-5290}, 3(01):17--22.

\bibitem[{Tran et~al.(2019)Tran, Tran, Nguyen, Nguyen, and Nguyen}]{tran2019does}
Ngoc Tran, Hieu Tran, Son Nguyen, Hoan Nguyen, and Tien Nguyen. 2019.
\newblock Does bleu score work for code migration?
\newblock In \emph{2019 IEEE/ACM 27th International Conference on Program Comprehension (ICPC)}, pages 165--176. IEEE.

\bibitem[{Wei et~al.()Wei, Wang, Schuurmans, Bosma, Xia, Chi, Le, Zhou et~al.}]{weichain}
Jason Wei, Xuezhi Wang, Dale Schuurmans, Maarten Bosma, Fei Xia, Ed~H Chi, Quoc~V Le, Denny Zhou, et~al.
\newblock Chain-of-thought prompting elicits reasoning in large language models.
\newblock In \emph{Advances in Neural Information Processing Systems}.

\bibitem[{Yin et~al.(2018)Yin, Deng, Chen, Vasilescu, and Neubig}]{yin2018mining}
Pengcheng Yin, Bowen Deng, Edgar Chen, Bogdan Vasilescu, and Graham Neubig. 2018.
\newblock \href {https://doi.org/https://doi.org/10.1145/3196398.3196408} {Learning to mine aligned code and natural language pairs from stack overflow}.
\newblock In \emph{International Conference on Mining Software Repositories}, MSR, pages 476--486. ACM.

\bibitem[{Zheng et~al.(2023)Zheng, Xia, Zou, Dong, Wang, Xue, Wang, Shen, Wang, Li et~al.}]{zheng2023codegeex}
Qinkai Zheng, Xiao Xia, Xu~Zou, Yuxiao Dong, Shan Wang, Yufei Xue, Zihan Wang, Lei Shen, Andi Wang, Yang Li, et~al. 2023.
\newblock Codegeex: A pre-trained model for code generation with multilingual evaluations on humaneval-x.
\newblock \emph{arXiv preprint arXiv:2303.17568}.

\bibitem[{Zhong et~al.(2022)Zhong, Liu, Yin, Mao, Jiao, Liu, Zhu, Ji, and Han}]{zhong2022towards}
Ming Zhong, Yang Liu, Da~Yin, Yuning Mao, Yizhu Jiao, Pengfei Liu, Chenguang Zhu, Heng Ji, and Jiawei Han. 2022.
\newblock Towards a unified multi-dimensional evaluator for text generation.
\newblock In \emph{Proceedings of the 2022 Conference on Empirical Methods in Natural Language Processing}, pages 2023--2038.

\bibitem[{Zhou et~al.(2023)Zhou, Alon, Agarwal, and Neubig}]{zhou2023codebertscore}
Shuyan Zhou, Uri Alon, Sumit Agarwal, and Graham Neubig. 2023.
\newblock Codebertscore: Evaluating code generation with pretrained models of code.
\newblock In \emph{Association for Computational Linguistics: EMNLP 2023}.

\bibitem[{Zhu et~al.(2023)Zhu, Wang, Zhou, Wang, Chen, Wang, Yang, Ye, Gong, Zhang et~al.}]{zhu2023promptbench}
Kaijie Zhu, Jindong Wang, Jiaheng Zhou, Zichen Wang, Hao Chen, Yidong Wang, Linyi Yang, Wei Ye, Neil~Zhenqiang Gong, Yue Zhang, et~al. 2023.
\newblock Promptbench: Towards evaluating the robustness of large language models on adversarial prompts.
\newblock \emph{arXiv preprint arXiv:2306.04528}.

\bibitem[{Zhuo et~al.(2023)Zhuo, Huang, Chen, and Xing}]{zhuo2023red}
Terry~Yue Zhuo, Yujin Huang, Chunyang Chen, and Zhenchang Xing. 2023.
\newblock \href {http://arxiv.org/abs/2301.12867} {Red teaming chatgpt via jailbreaking: Bias, robustness, reliability and toxicity}.

\end{thebibliography}

\appendix

\section{Prompts for Code Evaluation}

\subsection{Code Usefulness}
\label{appendix:usefulness}
\begin{quote}
\textit{Evaluation Criteria:}\\
\textit{Usefulness (0-4) Usefulness of the code snippet based on the problem description.}\\
\textit{- A score of 0: Snippet is not at all helpful, it is irrelevant to the problem.}\\
\textit{- A score of 1: Snippet is slightly helpful, it contains information relevant to the problem, but it is easier to write the solution from scratch.}\\
\textit{- A score of 2: Snippet is somewhat helpful, it requires significant changes (compared to the size of the snippet), but is still useful.}\\
\textit{- A score of 3: Snippet is helpful, but needs to be slightly changed to solve the problem.}\\
\textit{- A score of 4: Snippet is very helpful, it solves the problem.}
\end{quote}

\subsection{Functional Correctness}
\label{appendix:correctness}

\begin{quote}
\textit{Evaluation Criteria:}\\
\textit{Functional Correctness (0-4) - Execution-based quality of the code snippet combined with the problem. The correctness is measured by all possible unit tests and the comparison of the reference code. The combination of the code snippet and the problem should pass all the possible tests based on your understanding of the reference code. The length of the code snippet can not determine the correctness. You need to assess the logic line by line.}\\
\textit{- A score of 0  (failing all possible tests) means that the code snippet is totally incorrect and meaningless.}\\
\textit{- A score of 4  (passing all possible tests) means that the code snippet is totally correct and can handle all cases.}\\
\end{quote}

\section{Automatic Evaluation Metric Baselines}

Our implementations of the automatic evaluation metric baselines except for CodeBERTScore are based on \url{https://github.com/JetBrains-Research/codegen-metrics}. For CodeBERTScore, we adopt the official release at \url{https://github.com/neulab/code-bert-score}.

\section{Correlation Metrics}

For all correlation metrics, we use the implementation from \url{https://scipy.org/} and call these APIs with the default settings.

\section{Rule-based Score Extraction from Zero-shot Chain Of Thought Evaluation}

We demonstrate the general implementation of score extraction:

\begin{lstlisting}[language=Python, caption=Score Extractor Implementation]
import re
TASK_KEY_WORD = "usefulness" # or "functional"
def get_gpt_answer(raw_content):
    try:
        return int(raw_content)
    except:
        try:
            return process_raw_content(raw_content)
        except:
            return 0
            
def process_raw_content(content):
    # Clean up and split the content
    splits = content.lower().replace("(", "").replace(")", "").split("\n")

    # Extract relevant lines and clean them up
    ls = [ll.strip(".")
        .replace("out of ", "/")
        .replace("/4", "")
        for l in splits
        for ll in l.lstrip("0123456789. ").split(". ")
        if TASK_KEY_WORD in ll or "score" in ll]

    # Extract the scores
    ans = [ll for l in ls for ll in l.split() if ll.isnumeric()]

    # If there are multiple scores, take the most common one
    if len(set(ans)) != 1 and len(ans) > 1:
        return int(Counter(ans).most_common(1)[0][0])

    # If there are no scores or ambiguous scores, return 0 or 1
    if len(set(ans)) != 1:
        if "N/A" in content:
            return 0
        else:
            return 1

    # Otherwise, return the single score
    return int(ans[0])
\end{lstlisting}

We note that our extraction process for the evaluation metrics is entirely rule-based and may not be optimized for the best results.

\end{document}